\documentclass[letterpaper, 10 pt, conference]{ieeeconf}  
\makeatletter
\let\NAT@parse\undefined
\makeatother
\IEEEoverridecommandlockouts                              

\usepackage[numbers,sort&compress]{natbib}

\usepackage{epsfig}
\usepackage{graphicx}
\usepackage{amsmath}
\usepackage{amssymb}
\usepackage{booktabs}
\usepackage{times}
\usepackage{mathtools}
\usepackage{float}
\usepackage{xspace}
\usepackage{caption}
\usepackage{subcaption}
\usepackage[inline]{asymptote}
\usepackage{tikz}
\usepackage{makecell}
\usepackage{multirow}
\usepackage{multicol}
\usepackage{booktabs}
\usepackage{pifont}
\usepackage{pgfplots}
\usepackage{nomencl}
\usepackage[english]{babel}
\usepackage[autostyle]{csquotes}
\let\labelindent\relax

\usepackage[inline]{enumitem}
\usepackage{stackengine} 
\usepackage{resizegather}
\usepackage[accsupp]{axessibility}
\usepackage{hyperref}

\usepackage[utf8]{inputenc} 
\usepackage[T1]{fontenc}    
\usepackage{hyperref}       
\usepackage{url}            
\usepackage{booktabs}       
\usepackage{amsfonts}       
\usepackage{nicefrac}       
\usepackage{microtype}      
\usepackage{xcolor}         
\usepackage{graphicx}
\usepackage{caption}
\usepackage{amsmath}
\usepackage{amssymb}
\usepackage{mathtools}
\usepackage{wrapfig}
\usepackage{tabularx}

\usepackage[utf8]{inputenc} 
\usepackage[T1]{fontenc}    
\usepackage{hyperref}       
\usepackage{url}            
\usepackage{booktabs}       
\usepackage{amsfonts}       
\usepackage{nicefrac}       
\usepackage{microtype}      
\usepackage{multicol}
\usepackage{multirow}
\usepackage{adjustbox}
\usepackage{wrapfig}
\usepackage{changes}
\usepackage{xspace}
\usepackage{float}
\usepackage{cleveref}

\definechangesauthor[color=red]{SS}

\newcommand{\eg}[1]{\emph{e.g.,}\xspace}
\newcommand{\parahead}[1]{\noindent\textbf{#1}:\ }

\newcommand{\filluptopage}[1]{%
  \clearpage
  \loop\ifnum\value{page}<#1\relax
    \null\clearpage
  \repeat
  \loop\ifnum\value{page}=#1\relax
    \null\clearpage
  \repeat
}

\makeatletter
\def\blfootnote{\xdef\@thefnmark{}\@footnotetext}
\makeatother

\DeclareMathOperator*{\argmin}{arg\,min}

\newcommand{\cmark}{\textcolor{green}{\checkmark}}%
\newcommand{\xmark}{\textcolor{red}{\ding{55}}}%

\newcommand{\shortname}{MotionGlot\xspace}
\newcommand{\authorhref}[3][violet]{\href{#2}{\color{#1}{#3}}}
\newcommand{\papertitle}{A Multi-Embodied Motion Generation Model}

\newcommand{\datasetname}{QUAD-LOCO }
\newcommand{\captionname}{QUES-CAP }
\newcommand{\webpage}{https://ivl.cs.brown.edu/research/motionglot.html}
\title{\LARGE \bf \shortname: \papertitle \\
\small{\href{\webpage}{\color{magenta}{\webpage}}}  \vspace{-5mm} 
}

\author{ \authorhref{https://sudarshan-s-harithas.github.io/}{Sudarshan Harithas}$^1$, 
\authorhref{https://cs.brown.edu/people/ssrinath/}{Srinath Sridhar}$^1$ 
\thanks{$^1$Brown University, USA.}
\thanks{{\tt\small sudarshan\_harithas@brown.edu}}
}

\begin{document}

\thispagestyle{empty}
\pagestyle{empty}

\makeatletter
\let\@oldmaketitle\@maketitle
\renewcommand{\@maketitle}{\@oldmaketitle
\centering
\includegraphics[width=\textwidth]{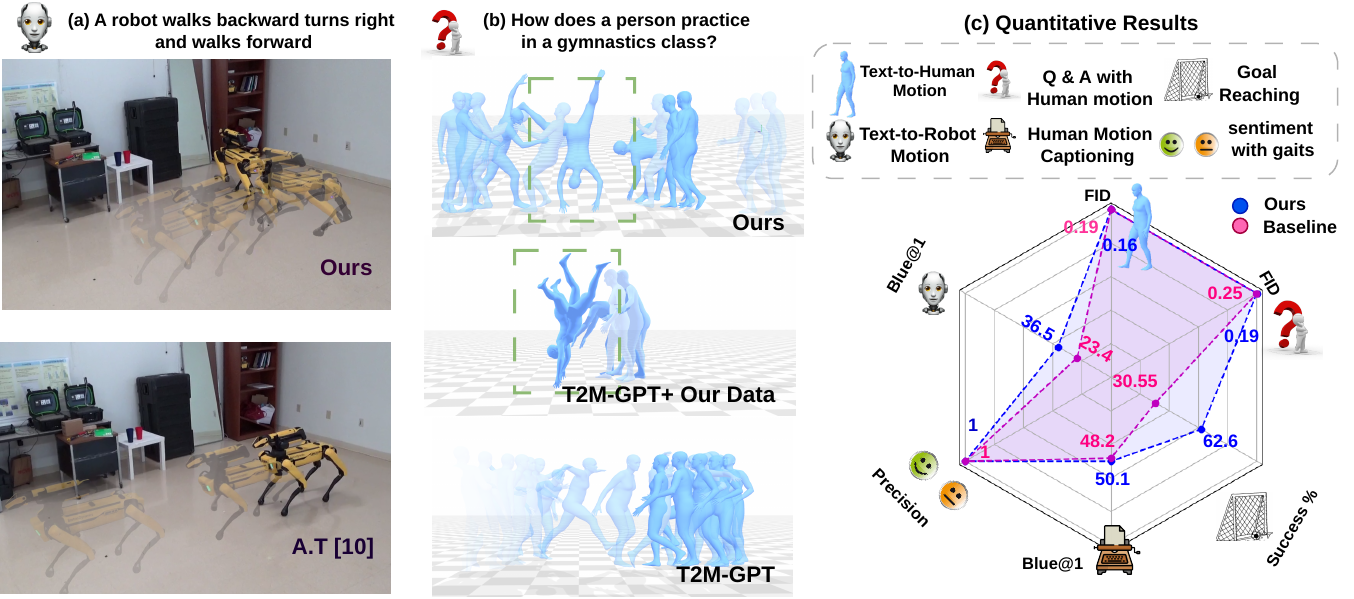}
\vspace{-5.5mm}
\captionof{figure}{\small{ \textbf{Overview}:  \shortname is a model that can generate motion trajectories that obey user instructions across multiple embodiments with different action dimensions, such as (a)~quadruped robots, and (b)~humans.
The figures (a,b) depict the qualitative benchmark of \shortname against the adapted templates \textbf{(A.T)} of \cite{rt2} on the text-to-robot motion (\Cref{sec:exp_t2rm}), Q\&A with human motion (\Cref{exp:q_and_a}) tasks respectively.
The overall quantitative performance across tasks is shown in (c).
In (a,b), increasing opacity indicates forward time. }
}
\label{teaser}
\vspace{-3.5mm}
}

\maketitle

\begin{abstract}

This paper introduces \shortname, a model that can generate motion across multiple embodiments with different action dimensions, such as quadruped robots and human bodies.
By leveraging the well-established training procedures commonly used in large language models (LLMs), we introduce an instruction-tuning template specifically designed for motion-related tasks.
Our approach demonstrates that the principles underlying LLM training can be successfully adapted to learn a wide range of motion generation tasks across multiple embodiments with different action dimensions.
We demonstrate the various abilities of \shortname on a set of \textbf{6} tasks and report an average improvement of 35.3\% across tasks. 
Additionally, we contribute two new datasets: (1)~a dataset of expert-controlled quadruped locomotion with approximately 48,000 trajectories paired with direction-based text annotations, and (2)~a dataset of over 23,000 situational text prompts for human motion generation tasks.
Finally, we conduct hardware experiments to validate the capabilities of our system in real-world applications.

\vspace{-3mm}

\end{abstract}

\section{INTRODUCTION}
Large Language Models (LLMs)~\cite{attention,gpt, gpt-4, gpt2, llama,llama3} have seen tremendous success recently with models that can produce text indistinguishable from human-generated text.
These models have also shown to be useful in applications beyond just text generation, for example, in multi-lingual translation~\cite{gpt2,mala}, multi-task learning~\cite{gpt2,gpt, gpt-4, gpt2, llama,llama3}, or instruction following~\cite{instruct_template}.

LLMs use transformers~\cite{attention} to model language as a sequence of tokens and are trained in a next-token or masked-token prediction framework.
Indeed, some research has looked into modeling other forms of sequential data using the same machinery, for example, in audio~\cite{wav2vec} and weather data~\cite{totem}.
Unsurprisingly, recent work has also modeled motion and action as a sequential generation problem~\cite{rt1,rt2,openvla}.
However, these approaches have thus far been limited to a single embodiment~\cite{t2mgpt,t2mt,motiongpt} or embodiments with the same number of action space dimensions~\cite{openvla,rt2}.

In this paper, we investigate the problem of building models of action that can cover multiple embodiments with different action spaces (\eg~humans vs.~quadrupeds).
This is a hard problem because (1)~motion data is not always plentifully available for all embodiments and (2) the action dimension and morphological constrains of operation widely varies across embodiments.  



We overcome these limitations with \textbf{\shortname}, a motion generation model that can span multiple embodiments with different action spaces. 
\shortname builds on top of the well-established instruction-tuning techniques from multilingual 
LLMs~\cite{mala,instruct_template,instruct_template2,instruction_tune} and proposes an instruction template to train a GPT~\cite{gpt2} for motion generation. While our insights and framework can be generalized and extended to multiple morphologies, we are primarily interested in two embodiments with different action spaces: \textbf{human bodies} and \textbf{quadruped robots}.
\shortname is a single model that exhibits core capabilities which are depecited in \Cref{teaser}, these include text-conditioned motion generation and motion captioning for multiple embodiments. 

To overcome the challenges of limited data availability for quadrupeds, we propose \datasetname, a dataset of expert-controlled quadruped locomotion with direction-based text annotation \Cref{fig:robot_imgs} (c). Additionally, we introduce a new dataset consisting of text captions for human motions. By harnessing the few-shot learning capabilities of \emph{GPT-4}~\cite{gpt-4}, we have generated over 23,000 situational descriptions of human actions. This dataset will be utilized for the Q\&A with human motion task (\Cref{exp:q_and_a}).

\datasetname not only enables our core capability such as text-conditioned locomotion for quadruped, but also additional capabilities such as goal-conditioned motion generation for quadrupeds.
Our experiments (\Cref{sec:exp}) demonstrate that \shortname is a generalist method that can generate motion across multiple embodiments, handle unseen user instructions, and express the multi-modal distribution in motion trajectories.
\shortname also performs better than existing methods as shown in \Cref{teaser} and \Cref{sec:exp}.

Overall, our contributions are:
\begin{enumerate*}[label=\textbf{(\arabic*)}]
    \item \shortname, a model that learns to generate motions across multiple embodiments with different action spaces. 
    \item an instruction tuning template that uses a single decoder-only transformer to generate motion across multiple embodiments and operate as a multi-task learner, and
    \item The \datasetname dataset which consists of \textbf{48000} quadruped trajectories with direction-based textual descriptions for robot motion and the \captionname dataset which consists of more than \textbf{23000} prompts that enable Q\&A with motion \Cref{exp:q_and_a}. 

 \end{enumerate*}

 \vspace{-3mm}
\section{Related Works}

In this brief review, we focus on the closest work in language, robotics, motion generation, and captioning.
Please see \Cref{tab:comparison_rel_work} for a summary of related works.

\parahead{Language and Robotics}
There has been an explosion of recent work at the intersection of language and robotic navigation or manipulation~\cite{cliport,conceptgraphs,vlmaps,rt1} that treat language as an additional modality and have separate branches in the network to process text instructions.

Methods such as RT-2~\cite{rt2} or OpenVLA~\cite{openvla} have attempted to unify language and action into a common vocabulary to train models for manipulation tasks. 
However, their instruction tuning template is largely limited to embodiments with the same action dimension (\eg~7DoF action space of a manipulator).
Driven by insights from multi-lingual instruction tuning~\cite{instruct_template,mala,instruct_template2,mgpt} our proposed method enables us to build a common vocabulary across embodiments with very different action spaces, specifically, human motions and quadruped motions.

Works such as ~\cite{gato,crossformer} leverage autoregressive transformers to create a common controller policy for multiple embodiments. Unlike these methods, \shortname serves a different objective and caters towards generative tasks. While RoboCat \cite{robotcat} attends towards building a common model across different output dimensions, their approach is demonstrated only on manipulators, whereas \shortname explores diverse embodiments such as quadrupeds and human bodies. 
Additionally, our proposed training procedures bring the instruction-following and multi-task learning abilities of LLMs into motion generators.

\parahead{Human and Robot Motion Generation}
Motion generation for human bodies and mobile robots has been largely studied in separate communities.
Human motion generation methods can be classified into two categories~\cite{survey}: (1)~methods that use pre-trained vision-language models like CLIP~\cite{clip} for motion generation~\cite{t2mgpt,momask,mdm,motiondiffuse}, and (2)~methods such as \cite{t2mt,motiongpt}, which jointly learn a text and motion representation. Works related to motion generation for robots have largely focused on works that have the same action dimensions such as \cite{vint,gnm,rt2,openvla}.While \shortname belongs to the second category, unlike all the aforementioned models, \shortname is a multi-embodied motion generator. 

\begin{table}[t]
\scriptsize
\vspace{3mm}
	\centering
	\setlength{\tabcolsep}{8pt}
		\begin{tabular}{lccccc}  
			\toprule
			\textbf{Method}  & \textbf{M-E} & \textbf{M-T} & \textbf{H/R M-G} &  \textbf{H/R M-C} \\
			\midrule
            
           Adapted templates of \cite{rt2,openvla} & \xmark & \cmark & \xmark / \cmark & \xmark / \xmark \\ 
           RoboCat \cite{robotcat} & \cmark & \cmark & \xmark / \cmark & \xmark / \xmark \\ 
                T2MGPT \cite{t2mgpt} & \xmark   & \xmark       & \cmark / \cmark   &  \xmark / \xmark \\ 
                T2MT \cite{t2mt}    & \xmark    & \xmark       & \cmark / \cmark    &  \cmark / \xmark \\
             MotionGPT \cite{motiongpt} & \xmark & \cmark       & \cmark / \xmark    & \cmark / \xmark  \\
         MDM \cite{motiondiffuse}    & \xmark    & \xmark      & \cmark / \xmark     & \xmark / \xmark \\
         \hline 
         Ours                         & \cmark   & \cmark      & \cmark / \cmark     & \cmark / \cmark \\
			\bottomrule
		\end{tabular} 	
	\caption{  \footnotesize{ Acronyms: \textbf{M-T}: Multi task ability, \textbf{H/R M-G}: Human/ Robot motion generation ability. \textbf{H/R M-C}: Human / Robot Motion captioning ability. \textit{Robot} refers to a quadruped robot whose locomotion can be controlled with \textit{SE2} velocity commands. \textbf{M-E} refer to the ability to perform generative tasks on multiple embodiements with different action dimensions. refer to Sec. ~\ref{sec:exp_t2rm} for adapted templates of \cite{openvla,rt2}. }
 } 
	\label{tab:comparison_rel_work}
 \vspace{-8mm}
\end{table}

\parahead{Datasets}
While there exist large pools of data for manipulation~\cite{openx} and navigation~\cite{diff_nav1,diff_nav2,diff_nav3}, there are no large data sources for quadruped locomotion paired with text.
While \cite{saytap} proposes to model quadruped gaits using their feet-floor contact pattern, the dataset largely ignores direction based annotation such as the captions shown in \Cref{fig:robot_imgs} (c). 
Therefore, to expand the text-conditioned motion generation capabilities to robots, we propose \datasetname, a dataset with over \textbf{48000} (after data-augmentation) pairs of expert-controlled real-world quadruped motion trajectories with direction-based text annotation (\Cref{sec:data_creation}).

For human body motion, the \textit{AMASS}~\cite{AMASS} dataset, which includes text annotations from \cite{humanml}, has been a key resource~\cite{momask,motiongpt,t2mgpt,t2mt,humanml}. While \cite{humanml} offers a broad range of action descriptions, it often lacks the contextual details of specific situations where these actions occur (~\Cref{sec:data_creation}).
To tackle this, we employed GPT-4~\cite{gpt-4} to enhance the descriptions from \cite{humanml} and generate \textbf{23,000} situation-based text descriptions, transforming them into questions (see \Cref{sec:data_creation}). This newly created dataset facilitates applications such as Q\&A with human motion tasks (see \Cref{exp:q_and_a}).

\parahead{Motion Captioning}
Motion captioning is the task of generating a text description for the input motion.
T2MT~\cite{t2mt} uses an Encoder-Decoder transformer to caption human motion, however, such approaches are constrained to a single task of bidirectional translation between text and motion.
MotionGPT~\cite{motiongpt} leverages a T5~\cite{t5} model for motion captioning and motion synthesis, however, \cite{motiongpt} is constrained to a single embodiment.
\cite{robot_motion_caption} performs captioning of robot actions, however, they are single-task, single embodiment models. In contrast, our model natively supports text captioning.

\section{Method}
\definecolor{maroon}{rgb}{0.6, 0, 0.6}

\renewcommand{\thefigure}{2}
\begin{figure*}
\centering
\vspace{2mm}
\includegraphics[width = \textwidth]{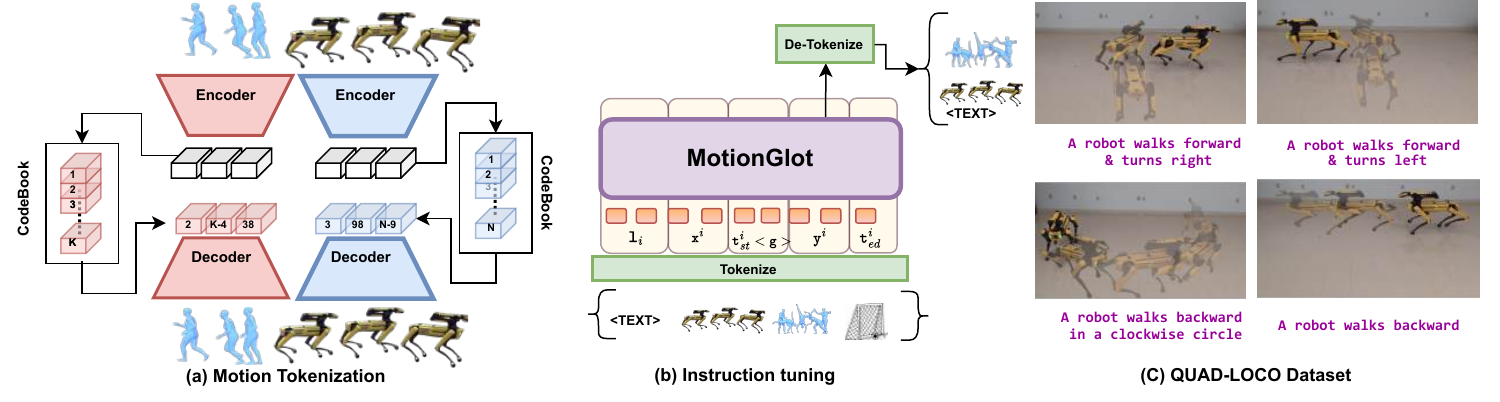}
	\caption{\footnotesize (a) Trajectories from different embodiments are tokenized using their associate VQ-VAE~\cite{vqvae} (\Cref{sec:tokenization}). (b) The proposed instruction template (\Cref{sec:motionglot_template}) is used to train GPT for motion and text generation. Note that the tokenizer and de-tokenizer operate on the expanded vocabulary \Cref{sec:vocab_exp} ($\mathcal{V}$)  (c) The preview of the \datasetname dataset, the \textcolor{maroon}{captions} indicate the direction-based text annotation.}
		\label{fig:robot_imgs}
\vspace{-7mm}
\end{figure*}

We intend to build a model capable of motion generation across multiple embodiments with different action spaces. We approach this problem as a next-token prediction problem similar to LLMs. \Cref{fig:robot_imgs} shows an overview of our approach.
Below, we describe individual components.  Our training procedure involves two steps, in the first stage a VQ-VAE~\cite{vqvae} learns a discrete latent codebook that represents a motion vocabulary per embodiment.  This process, known as motion tokenization, is similar to text tokenization~\cite{wordpiece}.
The motion vocabulary across embodiments are then appended to the existing vocabulary of GPT2~\cite{gpt} creating a unified motion and text vocabulary.
In the second step, our proposed instruction template is used to train the autoregressive GPT ~\cite{attention,gpt,gpt2}.
\vspace{-2mm}
\subsection{Trajectory Parameterization \& Tokenization} \label{sec:tokenization}
For a given embodiment, a motion trajectory of length $\mathcal{T}$ is parameterized as $\mathbf{x}^e = [p_0^e, p_1^e, \cdots, p_{\mathcal{T}}^e]$, where $p$ denotes motion represented as the embodiment's pose, and $e$ denotes different embodiments -- in our case either the quadruped robot ($r$) or human ($h$).
The quadruped trajectory is parameterized by a sequence of 2D linear ($\dot{x},\dot{z}$) and angular velocities ($\dot{r_{a}}$) where a pose at a discrete time $t$ is given by $p_t^r = ( \dot{x},\dot{z} , \dot{r_{a}} ) \in R^{SE2}$.
Here, we assume that the y-axis is perpendicular to the ground plane ($xz$).   
The human pose is parameterized using the canonical representation from SMPL~\cite{SMPL,humanml} as $ p_t^h = (\dot{r}_{a} , \dot{r}_{xz} , r_{y} , j_{p} ,j_{v} , j_{r} , c_{f} ) \in R^{263}$, where $\dot{r}_{xz} \in R^{2}$ is the root velocity along the ground plane, $\dot{r}_{a}$  $\in R^{1}$ is the root angular velocity along the y-axis, ${r}_{y} \in R^{1}$ is the height of root from ground, $j_{p}, j_{v} \in R^{3k}$ and $j_{r} \in R^{6k}$ refer joint positions, joint velocities and joint angles represented as continuous 6D vectors, and $c_{f} \in R^{4}$ are the foot contact features,  the number of joints $k=22$ for the \cite{humanml} dataset.

The goal of the tokenizer is to develop representations that allow a trajectory to be expressed as a series of discrete tokens, where each token is a unique element belonging to a finite vocabulary. We employ a \textit{VQ-VAE} \Cref{fig:robot_imgs} (a) \cite{vqvae} which consists of an autoencoder with a learnable codebook $\mathcal{C} \in R^{N \times d } $ with $N$ tokens each of embedding dimension $d$. 
A separate VQ-VAE \cite{vqvae} is maintained for each embodiment, where the codebook represents the learned vocabulary for that embodiment.

The motion trajectories ($\mathbf{x}_{e}$) are first passed through the encoder that applies 1D convolutions to create a latent code $z \in R^{d \times T/l }   $, where $l$ is the temporal down-sampling from the encoder. The quantization process substitutes each entry of the latent space $z_{i} \in R^{d}$ with the closest element in the codebook $\hat{z_{i}} \in R^{d}$ given by \Cref{eq:nearsest}.
The quantized embeddings $\hat{z_{i}}$, are then fed into the decoder to reconstruct the input signal $\hat{x} \in \mathbb{R}^{d_{e} \times T}$ as
\vspace{-2mm}
\begin{align}\label{eq:nearsest}
\vspace{-5mm}
    \hat{z_{i}} = \argmin _{c_{k} \in \mathcal{C}} || z_{i} - c_{k} ||_{2}.
\end{align}
\vspace{-3mm}

The tokenizer is trained using three loss functions \cite{vqvae,t2mgpt}: $L = L_{r} + L_{e} + L_{c}$, where $L_{r}$ represents the reconstruction loss, $L_{e}$ is the embedding loss, and $L_{c}$ denotes the commitment loss. Following the approach outlined in \cite{t2mgpt}, all loss functions are $L_{1}$ loss with smoothing, velocity regularization, and EMA with codebook reset techniques \cite{vqvae} are included.  

Note, that in contrast to discrete binning-based tokenization used in \cite{openvla,rt2} where $N$ tokens are used to represent a single pose of $N-DOF$ output space, using the \textit{VQ-VAE} based tokenization one token would return $l$ poses. Leading to a total compression of the order of $\mathcal{O}(lN)$, thereby, improving the use of the finite context window of the transformer~\cite{attention,gpt,llama}.

\vspace{-2mm}
\subsection{Instruction Tuning }\label{sec:motionglot_template}

To enable multi-embodiment motion synthesis we leverage insights from instruction tuning for multi-lingual models~\cite{instruct_template,instruct_template2,mala}. The process involves two steps, first, we merge the motion and text vocabularies to create a unified vocabulary suitable for generating motion and text. In the second step, we propose an instruction template for motion synthesis is proposed. We first define various vocabularies and their objectives.

\parahead{Vocabulary Definition}
\label{sec:vocab_define}
We choose GPT-2~\cite{gpt} as the backbone model for training, its vocabulary ($\mathcal{V}_{l}$) size of 50,257 primarily consists of tokens from the English language.
The VQ-VAE~\cite{vqvae} results in a motion vocabulary denoted as $\mathcal{V}_{r}, \mathcal{V}_{h}$ for the robot and human motion respectively.
Additionally, the ground plane is divided into uniform cells and each cell is treated as a token, the complete set of these cells forms the vocabulary $\mathcal{V}_{g}$. Furthermore, a vocabulary of gait tokens $\mathcal{V}_{gait}$ are defined that indicate the choice of gait the quadruped must choose while executing the trajectory, each of the gait tokens are associated with an RL-controller trained using proximal policy optimization (\textit{PPO}) \cite{ppo}, which execute the trajectory with the chosen gait. Following works from machine translation \cite{mala}, task-specific special tokens are included that indicate the start and end of the response, the vocabulary of special task identification tokens is given by $\mathcal{V}_{s}$.

\parahead{Vocabulary Expansion}
\label{sec:vocab_exp}
Following insights from instruction tuning strategies from multi-lingual LLMs~\cite{instruction,instruct_template,instruct_template2,instruction_tune}, we merge all the vocabularies, to create a single vocabulary given as $\mathcal{V} = \{ \mathcal{V}_{l}, \mathcal{V}_{r}, \mathcal{V}_{h}, \mathcal{V}_{s} , \mathcal{V}_{g} , \mathcal{V}_{gait} \}$. Performing next-token prediction on such a unified vocabulary ($\mathcal{V}$), across text, human, robot trajectories, and $2D$ ground plane enables the generation of motion across embodiments with different action dimensions in the same way text is generated.

\parahead{Training Template}  \label{sec:instruct_temp} Given a corpus $\mathcal{M}$ of input-output  ($\mathtt{x}^{i} ,\mathtt{y}^{i}$) pairs, a prefix ($l$) and the corresponding task-specific start ($t_{st}^{i}$) and end ($t_{ed}^{i}$) special tokens, the dataset is represented as $\mathcal{M} = \{ ( t_{st}^{i}, t_{ed}^{i}, \mathtt{x}^{i}, \mathtt{y}^{i} , l_{i} ) \} $. For a given sample $p_{i} \in \mathcal{M}$, we leverage a template $\hat{\mathcal{T}}$ to create a task instruction $d^{i}$, i.e. $d^{i} = \hat{\mathcal{T}}(p_{i})$.  The template $\hat{\mathcal{T}}$ is defined in Eq. ~\ref{eq:template}, where $<\mathtt{g}>$ is an optional field for the gait indicator token, which would only be active for robot trajectory generation. This stage is depicted in \Cref{fig:robot_imgs} (b).
%
\begin{align}
\label{eq:template}
\vspace{-6mm}
    \hat{\mathcal{T}} := l_{i}: \mathtt{x}^{i} \, t_{st}^{i} \, <\mathtt{g}> \,  \mathtt{y}^{i} \,  t_{ed}^{i}
\end{align}

Note that unlike the training strategies used \cite{rt2,openvla} our template is not restricted to a single embodiment. The standard next-token prediction objective from \cite{gpt,attention} on the vocabulary $\mathcal{V}$ is used to train the GPT.  The task-specific substitution for $l_{i}, \mathtt{x}_{i} , \mathtt{y}^{i} $ are detailed in Sec. ~\ref{sec:exp}.

\vspace{-2mm}

\subsection{Dataset Creation} \label{sec:data_creation}

\subsubsection{\textbf{\datasetname Dataset}}

Motion generation has largely been limited to single human embodiments due to the lack of data beyond human bodies ~\cite{AMASS,humanml,BABEL}. Therefore, we propose the \datasetname dataset with around $48000$ pairs (with data augmentation) of trajectories and direction-based text annotation. A preview of the \datasetname dataset is displayed in Fig. ~\ref{fig:robot_imgs} (c). Here, an expert operator remotely controls a spot quadruped robot to follow direction-based text-based instructions. The resulting movements of the robot are recorded, creating a dataset with quadruped motion and textual command correspondences. More than $1000$ trajectories have been recorded over $2.5$ hours from the expert teleoperator. Additionally, we apply the mirroring strategies from \cite{humanml} and time-scale the trajectories as further augmentation techniques. The \datasetname dataset has been crucial for enabling text-to-robot motion (Sec. ~\ref{sec:exp_t2rm}) and goal-conditioned motion generation (Sec. ~\ref{exp:goal_reach}).

\subsubsection{\textbf{\captionname Dataset}}

Datasets like \cite{humanml,BABEL} have advanced human motion generation, however, the captions typically lack the situational context in which the action can be performed.To enable human motion generators to synthesize motion based on situational queries, we propose the \captionname dataset. We leverage GPT-4's \cite{gpt-4} few-shot learning \cite{gpt3} capabilities to generate situational questions based on everyday scenarios and rewrite the provided text descriptions from \cite{humanml} to serve as potential answers. For example, for a description like \textit{'a person is boxing; they throw an uppercut, then dodge, and throw a few right jabs'}, a corresponding situational question might be \textit{'What sequence of movements describes a beginner learning basic boxing techniques?'}. Similarly, for a description like \textit{'a man raises his right arm, wiggles it, and then brings it back down'}, a relevant situational question could be \textit{'How would someone look if they were trying to get someone’s attention from across a noisy room using only their arm?'}. With similar examples we prompt \texttt{gpt-4-turbo} to rewrite $\mathbf{23000}$ prompts from \cite{humanml} as questions. This dataset has been used in the \textit{Q \& A} with human motion task (Sec. ~\ref{exp:q_and_a}).
\section{Experiments} \label{sec:exp}
We conduct experiments to specifically answer the following questions related to the generative abilities of \shortname:
\begin{enumerate}[label=\textbf{Q\arabic*}] 
\item Can the same machinery that is used to generate text be used to generate diverse motion across embodiments?
\item Can \shortname generalize to unseen user instructions? 
\item Can \shortname express multi-modal action distribution? 
\end{enumerate}

Experiments in Sec. ~\ref{exp:translation}, ~\ref{exp:q_and_a}, ~\ref{exp:sentiment} address \textbf{Q1} they are motion equivalent tasks of classical language problems.  Sec. ~\ref{sec:exp_t2rm}, and Sec. ~\ref{exp:goal_reach} answers \textbf{Q2}, \textbf{Q3} respectively.  

\parahead{Implementation Details}
We choose GPT-2 (small) ~~\cite{gpt} as our base model, the codebook size of the human motion tokenizer and robot motion tokenizer are $R^{512 \times 512 }$ and $R^{128 \times 512}$ respectively. For the goal-reaching task, we divide the $14m \times 14m $ ground plane into cells with a uniform resolution of $0.5 \times 0.5m$. 
The downsampling rate ($l$) of the VQ-VAE ~\cite{vqvae} is set to $4$ (($l=4$). Our model is trained on eight $NVIDIA- A5000$, for about $20k$ steps with a per-device batch size of $16$ and $4$ steps of gradient accumulation.  Adam optimizer ~\cite{diederik2014adam} with an initial learning rate of $5 \times 10^{-4}$, that decays with a cosine schedule has been used during training.

\parahead{Evaluation Metrics}
Evaluation protocols and procedures from \cite{humanml} have been used, global text and motion features are extracted to compute the metrics below. Pre-trained models ($\mathcal{M}_{h}$) and ($\mathcal{M}_{r}$) are motion feature extractors for human and robot motion, respectively.
($\mathcal{M}_{h}$) is pre-trained model from~\cite{humanml} and similarly we train another feature extractor  ($\mathcal{M}_{r}$) which produces  close features for matched text and robot-motion pairs, and vice versa. Furthermore, $95\%$ confidence is reported similar to \cite{humanml}.

\begin{enumerate*}[label=\textbf{(\arabic*)}]
    \item \textbf{Diversity (\textbf{Div}):} $N$ pairs are randomly sampled from a set of global-motion features and the average distance between them is computed. 
    \item \textbf{Multimodality}(\textbf{MMod}): For a given query $20$ motion samples are generated forming $10$ pairs of motion and the average distance between them is computed. 
    \item  \textbf{FID:} is the distribution distance between the features of generated and real motion ~\cite{FID}. 
    \item \textbf{Translation Metrics:} BERT-score ~\cite{bertscore} (BS),  Rouge ~\cite{rouge}, Cider ~\cite{cider}, Bleu@N~\cite{bleu} (B@N)  measure similarity between the ground truth and the generated text. 
    \item \textbf{Success \%:} $40$ trajectories are sampled per goal cell and a trajectory is successful if it terminates within the target cell. 
    \item  \textbf{R-precision (RP):} For every generated output $\hat{\mathtt{y}}$, $32$ input conditions (either text or motion)  are sampled $\{\Tilde{\mathtt{x}}\}_{i=1}^{32}$ (1 ground truth and $31$ randomly sampled from dataset). The Euclidian distance between the features of $\hat{\mathtt{y}}$ and $\{\mathtt{x}\}_{i=1}^{32}$ are ranked to measure the retrieval accuracy. 
\end{enumerate*}

\subsection{ Translation } \label{exp:translation}

\subsubsection{\textbf{Text-to-Robot Motion}} \label{sec:exp_t2rm}
This experiment evaluates the ability of \shortname to follow unseen user instructions, the task is to generate trajectories that semantically follow the input direction-based text description from the test \datasetname dataset. While ~\cite{rt2,openvla} are primarily meant for manipulation tasks, here we adapt their instruction template to perform text-to-robot motion generation. Here we briefly detail the performed modifications to \cite{rt2}. Following ~\cite{openvla} the data has been cleaned from outliers by selecting samples between $1^{st}$ and $99$ quantiles. Each of the continuous dimensions has been uniformly discretized into $256$ bins, where each bin represents an action token. The target for the \textit{LLM} is obtained by concatenating the action tokens for each dimension with a space character as given in Eq. ~\ref{eq:openvla_template}. The string is given below where $\Delta x, \Delta y, \Delta \psi$ represent the $2D$ linear and angular velocities. ~\cite{openvla,rt2} further requires observation as the input, here we project the global $SE(2)$ position through a linear layer to serve as the observation. 

\vspace{-4mm}

\begin{equation}\label{eq:openvla_template}
\vspace{-3mm}
    terminate \Delta x \Delta y \Delta \psi
\end{equation}

The performance results are summarized in Table. ~\ref{tab:t2rm}. To quantitatively evaluate the performance in the text-to-robot motion task, we translate the input text instruction to a robot motion and back-translate the resulting motion tokens to get the text caption (refer to Sec. ~\ref{sec:rm2t} for the evaluation of the robot motion captioning ability), the metrics \textbf{B@4, B@1 and BS} are then used to measure the cycle consistency between the user text instruction and back-translation. Text and motion feature vectors from $\mathcal{M}_{r}$, are used in the measurement of \textbf{RP}.  A higher value of these metrics indicates greater consistency and adherence to the input text instruction. \textbf{Div} and \textbf{MMod} are used to evaluate the generative abilities of the model

For this task \texttt{"give robot motion: "} is substituted as the prefix $l_{i}$ in Eq. ~\ref{eq:template}, similarly, $\mathtt{x}_{i}$ is the sequence of text tokens and $\mathtt{y}^{i}$ is the sequence of robot motion tokens. \shortname outperforms competitors by  31.2\% on average across all back translation metrics. The qualitative results are shown in \Cref{teaser} (a), it can be observed while \shortname follows the user instructions, the adapted version of \cite{openvla,rt2} only execute the backward motion and does not turn right and walk forward.

\begin{table*}[h!]
\centering
\vspace{3mm}
\begin{tabular}{lcccccc}
\hline 
\textbf{Method} & $\textbf{B@4}\uparrow$  & $\textbf{B@1}\uparrow$  & $\textbf{BS}\uparrow$  & $\textbf{RP@1/2/3}\uparrow$  & $\textbf{Div}\rightarrow$ & $\textbf{MMod}\uparrow$  \\
\hline
Real & - & - & - & $0.26/0.47/0.579^{ \pm .001}$                                              & $4.10^{ \pm .003}$ & - \\
Ours   & $\mathbf{36.5^{ \pm .002}}$  & $\mathbf{64.7}^{ \pm .002}$ & $\mathbf{ 57.5}^{ \pm .003}$ & $\mathbf{0.18/0.35/0.48}^{ \pm .005} $ & $\mathbf{3.74}^{ \pm .011}$ & $2.35^{ \pm .022}$ \\
~\cite{rt2}$^{\mathbf{A.T}}$    & $23.4^{ \pm .003}$ & $51.1^{ \pm .002}$ & $35.9^{ \pm .003}$ & $0.045/0.095/0.156^{ \pm .002}$ & $3.35^{ \pm .012}$ & $\mathbf{3.18}^{ \pm .015}$ \\ 

\hline 
\end{tabular}
\vspace{-1mm}

\caption{  \footnotesize{ Results on the \datasetname test set. \textbf{A.T}: Adapted templates. $\uparrow$, $\downarrow$ indicate higher, lower the better respectively and $\rightarrow$ indicates closer to the real value the better. \textbf{Bold} indicates the best method, $\pm$ indicates the $95\%$ confidence interval as \cite{humanml} defines. } }
\label{tab:t2rm}
\vspace{-4mm}
\end{table*}

\subsubsection{ \textbf{Text-to-Human Motion} } \label{sec:exp_t2hm}

We evaluate the model's ability to generate motion across various embodiments with different action dimensions by conducting text-to-human motion on the test set of ~\cite{humanml}. text-to-human motion generation literature falls into two main categories. The first category (\texttt{Cat I}) includes methods such as ~\cite{t2mgpt,momask,mdm,motiondiffuse}, which use \textit{CLIP} \cite{clip} embeddings for motion generation. Techniques \cite{mdm,momask}, also use privileged information, such as ground-truth trajectory length, during evaluation.

The second category (\texttt{Cat II}) consists of methods like \shortname and ~\cite{t2mt,motiongpt} which don't use privilege information like \textit{CLIP} or ground-truth trajectory length, instead jointly learn both the text and motion representations. While \texttt{Cat I} are better than \texttt{Cat II} on metrics like \textit{FID, R-Precision,} and \textit{MMDist}, they are single-task specialized models. Conversely, \texttt{Cat II} methods offer greater versatility but trade-offs some performance in favor of their multi-tasking capabilities.

For this task, \texttt{"give human motion: "} is substituted as the task specific prefix $l_{i}$ in Eq. ~\ref{eq:template}, $x_{i}$ and $y_{i}$ are seuqnce of text and human motion tokens respectively.  
Tab. ~\ref{tab:t2hm_table} summarizes the performance in text-human motion task.  Where we compare to methods within \texttt{Cat II}, as they are directly comparable when privileged information is not used, however, Tab ~\ref{tab:t2hm_table} mentions \texttt{Cat I} for completeness. \shortname demonstrates a competitive performance against competing SOTA baselines.

\begin{table*}[h!]
    \centering
    \begin{tabular}{l c ccc ccc c}
        \toprule
      \textbf{Txt.Rep} & \textbf{Methods} & \multicolumn{3}{c}{\textbf{RPrecision$\uparrow$}} & \textbf{FID$\downarrow$} & \textbf{MMDist$\downarrow$} & \textbf{Diversity$\rightarrow$} & \textbf{MMod$\uparrow$}  \\
        \cmidrule(lr){3-5}
        & & \textbf{Top1} & \textbf{Top2} & \textbf{Top3} &  &  &  &   \\
        \midrule
       & Real & $0.511^{ \pm .003}$& $0.703^{ \pm .003}$ & $0.797^{ \pm .002}$ & $0.002^{ \pm .000}$ & $2.974^{ \pm .008}$ & $9.503^{ \pm .065}$ & -  \\

        \hline 

      &  MDM ~\cite{mdm}$^{\Delta}$ & $0.32^{ \pm .005}$ & $0.498^{ \pm .004}$ & $0.611^{ \pm .007}$ & $0.544^{ \pm .044}$ & $5.566^{ \pm .027}$ & \textbf{$9.559^{ \pm .086}$} & $2.799^{ \pm .072}$   \\
  \textbf{\texttt{Cat I}}    &  T2M-GPT ~\cite{t2mgpt} & $0.491^{ \pm .003}$ & $0.680^{ \pm .003}$ &$ 0.775^{ \pm .002}$ &  $0.116^{ \pm .004}$ & $3.118^{ \pm .011}$ & $9.761^{ \pm .081}$ & $1.856^{ \pm .011}$  \\
       &  MO-MASK ~\cite{momask}$^{\Delta}$ & $0.521^{ \pm .002}$ & $0.713^{ \pm .002}$ & $0.807^{ \pm .002}$ & $0.045^{ \pm .002}$ &  $2.958^{ \pm .008}$ & - &  $1.241^{ \pm .040}$   \\ 
        \hline \\

      &  T2MT ~\cite{t2mt} & $\mathbf{0.424^{ \pm .003}}$ & $\mathbf{0.618^{ \pm .003}}$ & $\mathbf{0.729^{ \pm .002}}$ & $1.501^{ \pm .017}$ & $\mathbf{3.467^{ \pm .011}}$ & $8.589^{ \pm .076}$ & $2.424^{ \pm .093}$   \\

 \textbf{\texttt{Cat II}}    &  MotionGPT ~\cite{motiongpt}$^{\delta_{1}}$ & $0.402^{ \pm .003}$ & $0.567^{ \pm .002}$ & $0.649^{ \pm .002}$ & \underline{$0.19^{ \pm .0056}$} & $4.18^{^{ \pm .001}}$ & $\mathbf{9.33^{ \pm .008}}$ & \underline{$3.43^{ \pm .11}$}\\

      &  Ours & \underline{$0.406^{ \pm .005} $} & \underline{$0.571^{ \pm .007}$} &  \underline{ $0.652^{ \pm .007} $} & $ \mathbf{ 0.1618^{ \pm .005}}$ & \underline{$3.969^{\pm .008}$} &   \underline{$9.724^{\pm .065}$}  & $\textbf{3.48}^{\pm .098}$ \\ \\
        \hline 
    \end{tabular}
    \caption{  \footnotesize{ Text to Human Motion Benchmark on the HumanML3D ~\cite{humanml} dataset. $\Delta$ indicates results evaluated with ground truth motion length. All values for the baselines are extracted from the paper, apart from $\delta_{1}$ which is from the pre-trained open source model. \underline{underline} is the second best method. Real data is deterministic therefore MMod is "-", and the Diversity value of \cite{momask} is not available. } }
    \label{tab:t2hm_table}
    \vspace{-2mm}
\end{table*}

\subsubsection{\textbf{Motion Captioning} }

\begin{table*}[h]
\centering
\begin{tabular}{lcc ccccccccc}
\toprule
\textbf{Embodiment}& \textbf{Methods} & \hspace{2mm} \textbf{RPrecision$\uparrow$} \hspace{-13mm} & & \textbf{MMDist$\downarrow$} & \textbf{Length$_{avg}\uparrow$} & \textbf{Bleu@1$\uparrow$} & \textbf{Bleu@4$\uparrow$} & \textbf{Rouge$\uparrow$} & \textbf{Cider$\uparrow$} & \textbf{BertScore$\uparrow$} \\
\cmidrule(lr){3-4}

&  & \textbf{Top1} & \textbf{Top3} & & & & & & & \\
\midrule
& Real & 0.523 & \textbf{0.828} & 2.901 & 12.75 & - & - & - & - & - \\
& TM2T ~\cite{t2mt} & \underline{ 0.516} & 0.823 & 2.935 & 10.67 & 48.9 & 7.00 & \underline{38.1} & 16.8 & 0.32 \\
\textbf{Human} & MotionGPT ~\cite{motiongpt} & \textbf{0.543} & \underline{0.827} & \underline{2.821} & \underline{13.04} & \underline{48.2} & \underline{12.47} & 37.4 & \underline{29.2} & \underline{0.324} \\
& Ours  & 0.508  & 0.805 & \textbf{2.78} & \textbf{14.42} &\textbf{ 50.1 }& \textbf{13.5} & \textbf{41.8} & \textbf{33.6} &  \textbf{0.339} \\

\bottomrule
\end{tabular}
\caption{ \footnotesize Motion Captioning Benchmark on HumanML3D ~\cite{humanml} dataset.} 
\label{tab:hm2t}
\vspace{-6mm}
\end{table*}

This task involves generating a text description for the input motion trajectory, the experiment further demonstrates the multi-task learning ability of \shortname, the results are given in \Cref{tab:hm2t}. The task-specific prefix  $l_{i}$ in Eq. ~\ref{eq:template} is \texttt{"give text description: }, $\mathtt{x}_{i}$ is the sequence of human motion tokens and $\mathtt{y}_{i}$ is the sequence of text tokens. We evaluate the performance of \shortname against the current SOTA human motion captioning techniques, our method delivers an average improvement of 6.5 \% on the motion captioning tasks across \textit{Bleu \cite{bleu}, Cider \cite{cider} and BertScore \cite{bertscore} }. The results indicate that the captions generated by \shortname accurately capture the input motion trajectory.

\vspace{-2mm}
\subsection{\textbf{Goal conditioned Motion Generation}} \label{exp:goal_reach} This experiment evaluates the ability of the model to express multi-modal action distributions, the task is to generate diverse trajectories that approach the goal. The task-specific prefix in Eq. ~\ref{eq:template} is $l_{i}$ is \texttt{"reach goal: "}, the input token $\mathtt{x}_{i}$ is the goal cell token from $\mathcal{V}_{g}$ an the output $\mathtt{y}_{i}$ is the robot motion tokens. The qualitative and quantitative results are shown in Fig. ~\ref{fig:goal_reach} and Tab. ~\ref{tab:goal_reach} respectively. A success is defined if the terminal position of the trajectory is within the goal cell. Diffusion with classifier guidance ~\cite{diffuser} is a promising generative approach for capturing multiple behavioral modes in the trajectory distribution, so we use it as a baseline for the goal-reaching task, training it on the \datasetname dataset. \shortname shows significant improvement over ~\cite{diffuser} in success 

\begin{table}[h!]
\centering
    \resizebox{\columnwidth}{!}{%
\begin{tabular}{ccccc}
\hline
\textbf{Method } & \textbf{Success $\uparrow$ \%} & \textbf{Diversity$\rightarrow$}  & \textbf{FID$\downarrow$}  & \textbf{MMod$\rightarrow$}  \\
\hline 
Real & 100 & $2.85^{\pm 0.031}$  & $0.039^{\pm 0.00}$  & $1.38^{\pm 0.0067}$ \\ 
Ours & $\mathbf{62.0}^{\pm 0.061}$ & $\mathbf{3.24}^{\pm 0.16}$  &  $\mathbf{0.33}^{\pm 0.014} $ & $\mathbf{1.56}^{\pm 0.01}$    \\
Diffusion ~\cite{diffuser} & $30.55^{\pm 0.074}$   & $3.51^{\pm 0.0106}$ & $0.95^{\pm 0.022}$ & $2.91^{\pm 0.009}$   \\ 
\hline
\end{tabular}
}
\caption{ \footnotesize Goal reaching Task. }
\vspace{-4mm}
\label{tab:goal_reach}
\end{table}

\vspace{-2mm}

\subsection{\textbf{ Q\&A with Human Motion} } \label{exp:q_and_a}


This task generates motion in response to user questions. Qualitative and quantitative results are shown in Fig. ~\ref{teaser} (b) and Tab. ~\ref{tab:q_a}. Fig. ~\ref{teaser} (b) shows that the motion from \cite{t2mgpt} is a generic walking motion, unrelated to gymnastics practice. After training on the \captionname dataset, the response improves, performing a headstand. Motion from \shortname is more expressive, like a cartwheel, aligning with the query. These results show \captionname can train models for motion-based \textit{Q \& A}. Performance improvements are summarized in Tab. ~\ref{tab:q_a}, with Eq. ~\ref{eq:template} matching Sec. ~\ref{sec:exp_t2hm}.


\begin{table}[h!]
    \resizebox{\columnwidth}{!}{%
    \begin{tabular}{lc ccc c}
        \toprule
        \textbf{Methods} & {\textbf{RP@3$\uparrow$}} & \textbf{FID$\downarrow$}  & \textbf{Div$\rightarrow$} & \textbf{MMod$\uparrow$} \\
        \midrule
        Real & $0.364^{ \pm .002}$ & $0.002^{\pm .000}$ &  $9.503^{\pm .065}$ & - \\

        T2M-GPT  & $\mathbf{0.38^{ \pm .003}}$ & $3.5^{ \pm .008}$ &  $8.58^{ \pm .078}$ & $2.89^{ \pm .042}$ \\
        T2m-GPT*  & $0.33^{ \pm .006}$ & $0.25^{ \pm .005}$ & $9.26^{ \pm .071}$ & $2.44^{ \pm .053}$ \\

        Ours & \underline{$0.36^{ \pm .003}$}   & $\mathbf{0.19}^{ \pm .006}$ & $\mathbf{9.69}^{ \pm .08}$ & $\mathbf{3.06}^{ \pm .042}$ \\ \\

 \hline
    \end{tabular}
    }
    \caption{ \footnotesize Q\& A with Motion. T2M-GPT* indicates \cite{t2mgpt} trained with \cite{humanml} and \captionname datasets.  }
    \label{tab:q_a}
    \vspace{-6mm}
\end{table}

\subsection{ \textbf{Ablation Studies} }

\subsubsection{\textbf{Robot Motion Captioning}}\label{sec:rm2t}

This ablation aims to generate direction-based text captions for robot trajectories. In Eq. ~\ref{eq:template}, $\mathtt{x}_{i}$ and $\mathtt{y}_{i}$ are the sequence of robot motion and text tokens respectively. The performance analysis is given in Tab, ~\ref{tab:rm2t}, the high value of translation metrics indicate that \shortname is a reliable motion-to-text translator. 

\begin{table}[h]
\centering
\resizebox{\columnwidth}{!}{%
\begin{tabular}{lcccccccccc}
\toprule
 \textbf{Methods} & \textbf{RP@3$\uparrow$} &  \textbf{MDist$\downarrow$} & \textbf{L$_{avg}\uparrow$} & \textbf{B@1$\uparrow$} & \textbf{B@4$\uparrow$} & \textbf{\cite{rouge}$\uparrow$} & \textbf{\cite{cider}$\uparrow$} & \textbf{\cite{bertscore}$\uparrow$} \\

\hline \\
 {Real} &  0.581  &  3.9 & 9.26 &  -  & -& - & - & - \\

 Ours  & 0.2635  & 3.09 & 8.58  &  64.7 & 41.1 & 74.5 & 29.6  &  0.6165 \\
 
\bottomrule

\end{tabular}
}
\caption{ \footnotesize Motion Captioning ablation on \datasetname dataset.} 
\label{tab:rm2t} \label{tab:exp_rm2t}
\vspace{-3mm}
\end{table}

\subsubsection{\textbf{Sentiment Classification with Gaits}} \label{exp:sentiment}

\cite{saytap} demonstrates that each sentiment class can be associated with a gait for robot locomotion. For example, the \textit{bounding} and \textit{trott} gait can be used to indicate happy and neutral sentiments. With \shortname the gait field in Eq. ~\ref{eq:template} indicates the sentiment, for instance, given an instruction "\textit{a robot joyfully walks forward}" the gait indicator token $<g>$ is set to the bounding gait, whereas for the instruction "\textit{a robot walks forward}" $<g>$ is set to trotting gait. $100$ samples from the \datasetname dataset was used to benchmark against GPT-4 ~\cite{gpt-4} in a few shot setting. The average precision of both methods is $100 \%$.

\renewcommand{\thefigure}{3}
 \begin{figure}
\centering
\includegraphics[width = \columnwidth, height=0.4\columnwidth]{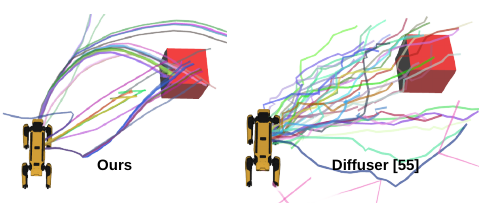}
  \vspace{-8mm}

 \caption{\footnotesize Qualitative results of the goal reaching task: note that our method expresses the multi-modal nature of the trajectory distribution, while \cite{diffuser} generates path towards the goal, its  success of convergence at goal is lower. }
		\label{fig:goal_reach}
  \vspace{-5mm}
\end{figure}

\vspace{-1mm}

\section{Conclusion}
\vspace{-1mm}
We introduce \shortname, a motion generator adaptable across embodiments with varying action space dimensions. Inspired by multilingual LLM training strategies, we propose a unified training template for motion generation. Our results show that \shortname generalizes to unseen user instructions, captures multi-modal action distributions, and functions as a multi-task learner across motion and text data.

\section*{Acknowledgements}

This research was supported by the Office of Naval Research (ONR) grant N00014-22-1-259.

\footnotesize{
\bibliographystyle{IEEEtran}
\bibliography{references}

\begin{thebibliography}{10}
\providecommand{\url}[1]{#1}
\csname url@samestyle\endcsname
\providecommand{\newblock}{\relax}
\providecommand{\bibinfo}[2]{#2}
\providecommand{\BIBentrySTDinterwordspacing}{\spaceskip=0pt\relax}
\providecommand{\BIBentryALTinterwordstretchfactor}{4}
\providecommand{\BIBentryALTinterwordspacing}{\spaceskip=\fontdimen2\font plus
\BIBentryALTinterwordstretchfactor\fontdimen3\font minus \fontdimen4\font\relax}
\providecommand{\BIBforeignlanguage}[2]{{%
\expandafter\ifx\csname l@#1\endcsname\relax
\typeout{** WARNING: IEEEtran.bst: No hyphenation pattern has been}%
\typeout{** loaded for the language `#1'. Using the pattern for}%
\typeout{** the default language instead.}%
\else
\language=\csname l@#1\endcsname
\fi
#2}}
\providecommand{\BIBdecl}{\relax}
\BIBdecl

\bibitem{rt2}
A.~Brohan, N.~Brown, J.~Carbajal, Y.~Chebotar, X.~Chen, K.~Choromanski, T.~Ding, D.~Driess, A.~Dubey, C.~Finn, P.~Florence, C.~Fu, M.~G. Arenas, K.~Gopalakrishnan, K.~Han, K.~Hausman, A.~Herzog, J.~Hsu, B.~Ichter, A.~Irpan, N.~Joshi, R.~Julian, D.~Kalashnikov, Y.~Kuang, I.~Leal, L.~Lee, T.-W.~E. Lee, S.~Levine, Y.~Lu, H.~Michalewski, I.~Mordatch, K.~Pertsch, K.~Rao, K.~Reymann, M.~Ryoo, G.~Salazar, P.~Sanketi, P.~Sermanet, J.~Singh, A.~Singh, R.~Soricut, H.~Tran, V.~Vanhoucke, Q.~Vuong, A.~Wahid, S.~Welker, P.~Wohlhart, J.~Wu, F.~Xia, T.~Xiao, P.~Xu, S.~Xu, T.~Yu, and B.~Zitkovich, ``Rt-2: Vision-language-action models transfer web knowledge to robotic control,'' in \emph{arXiv preprint arXiv:2307.15818}, 2023.

\bibitem{attention}
A.~Vaswani, N.~Shazeer, N.~Parmar, J.~Uszkoreit, L.~Jones, A.~N. Gomez, {\L}.~Kaiser, and I.~Polosukhin, ``Attention is all you need,'' \emph{Advances in neural information processing systems}, vol.~30, 2017.

\bibitem{gpt}
B.~Mann, N.~Ryder, M.~Subbiah, J.~Kaplan, P.~Dhariwal, A.~Neelakantan, P.~Shyam, G.~Sastry, A.~Askell, S.~Agarwal \emph{et~al.}, ``Language models are few-shot learners,'' \emph{arXiv preprint arXiv:2005.14165}, 2020.

\bibitem{gpt-4}
J.~Achiam, S.~Adler, S.~Agarwal, L.~Ahmad, I.~Akkaya, F.~L. Aleman, D.~Almeida, J.~Altenschmidt, S.~Altman, S.~Anadkat \emph{et~al.}, ``Gpt-4 technical report,'' \emph{arXiv preprint arXiv:2303.08774}, 2023.

\bibitem{gpt2}
A.~Radford, J.~Wu, R.~Child, D.~Luan, D.~Amodei, and I.~Sutskever, ``Language models are unsupervised multitask learners,'' 2019.

\bibitem{llama}
H.~Touvron, T.~Lavril, G.~Izacard, X.~Martinet, M.-A. Lachaux, T.~Lacroix, B.~Rozi{\`e}re, N.~Goyal, E.~Hambro, F.~Azhar \emph{et~al.}, ``Llama: Open and efficient foundation language models,'' \emph{arXiv preprint arXiv:2302.13971}, 2023.

\bibitem{llama3}
A.~Dubey, A.~Jauhri, A.~Pandey, A.~Kadian, A.~Al-Dahle, A.~Letman, A.~Mathur, A.~Schelten, A.~Yang, A.~Fan \emph{et~al.}, ``The llama 3 herd of models,'' \emph{arXiv preprint arXiv:2407.21783}, 2024.

\bibitem{mala}
P.~Lin, S.~Ji, J.~Tiedemann, A.~F. Martins, and H.~Sch{\"u}tze, ``Mala-500: Massive language adaptation of large language models,'' \emph{arXiv preprint arXiv:2401.13303}, 2024.

\bibitem{instruct_template}
J.~Li, H.~Zhou, S.~Huang, S.~Cheng, and J.~Chen, ``Eliciting the translation ability of large language models via multilingual finetuning with translation instructions,'' \emph{Transactions of the Association for Computational Linguistics}, vol.~12, pp. 576--592, 2024.

\bibitem{wav2vec}
\BIBentryALTinterwordspacing
A.~Baevski, Y.~Zhou, A.~Mohamed, and M.~Auli, ``wav2vec 2.0: A framework for self-supervised learning of speech representations,'' in \emph{Advances in Neural Information Processing Systems}, H.~Larochelle, M.~Ranzato, R.~Hadsell, M.~Balcan, and H.~Lin, Eds., vol.~33.\hskip 1em plus 0.5em minus 0.4em\relax Curran Associates, Inc., 2020, pp. 12\,449--12\,460. [Online]. Available: \url{https://proceedings.neurips.cc/paper_files/paper/2020/file/92d1e1eb1cd6f9fba3227870bb6d7f07-Paper.pdf}
\BIBentrySTDinterwordspacing

\bibitem{totem}
S.~Talukder, Y.~Yue, and G.~Gkioxari, ``Totem: Tokenized time series embeddings for general time series analysis,'' \emph{arXiv preprint arXiv:2402.16412}, 2024.

\bibitem{rt1}
A.~Padalkar, A.~Pooley, A.~Jain, A.~Bewley, A.~Herzog, A.~Irpan, A.~Khazatsky, A.~Rai, A.~Singh, A.~Brohan \emph{et~al.}, ``Open x-embodiment: Robotic learning datasets and rt-x models,'' \emph{arXiv preprint arXiv:2310.08864}, 2023.

\bibitem{openvla}
M.~J. Kim, K.~Pertsch, S.~Karamcheti, T.~Xiao, A.~Balakrishna, S.~Nair, R.~Rafailov, E.~Foster, G.~Lam, P.~Sanketi \emph{et~al.}, ``Openvla: An open-source vision-language-action model,'' \emph{arXiv preprint arXiv:2406.09246}, 2024.

\bibitem{t2mgpt}
J.~Zhang, Y.~Zhang, X.~Cun, S.~Huang, Y.~Zhang, H.~Zhao, H.~Lu, and X.~Shen, ``T2m-gpt: Generating human motion from textual descriptions with discrete representations,'' \emph{arXiv preprint arXiv:2301.06052}, 2023.

\bibitem{t2mt}
C.~Guo, X.~Zuo, S.~Wang, and L.~Cheng, ``Tm2t: Stochastic and tokenized modeling for the reciprocal generation of 3d human motions and texts,'' in \emph{European Conference on Computer Vision}.\hskip 1em plus 0.5em minus 0.4em\relax Springer, 2022, pp. 580--597.

\bibitem{motiongpt}
B.~Jiang, X.~Chen, W.~Liu, J.~Yu, G.~Yu, and T.~Chen, ``Motiongpt: Human motion as a foreign language,'' \emph{Advances in Neural Information Processing Systems}, vol.~36, 2024.

\bibitem{instruct_template2}
\BIBentryALTinterwordspacing
{Anoop Kunchukuttan}. Extending english large language models to new languages: A survey. [Online]. Available: \url{https://anoopkunchukuttan.gitlab.io/publications/presentations/ex tend_en_llms_apr2024.pdf}
\BIBentrySTDinterwordspacing

\bibitem{instruction_tune}
S.~Mishra, D.~Khashabi, C.~Baral, and H.~Hajishirzi, ``Cross-task generalization via natural language crowdsourcing instructions,'' \emph{arXiv preprint arXiv:2104.08773}, 2021.

\bibitem{cliport}
M.~Shridhar, L.~Manuelli, and D.~Fox, ``Cliport: What and where pathways for robotic manipulation,'' in \emph{Conference on robot learning}.\hskip 1em plus 0.5em minus 0.4em\relax PMLR, 2022, pp. 894--906.

\bibitem{conceptgraphs}
Q.~Gu, A.~Kuwajerwala, S.~Morin, K.~M. Jatavallabhula, B.~Sen, A.~Agarwal, C.~Rivera, W.~Paul, K.~Ellis, R.~Chellappa \emph{et~al.}, ``Conceptgraphs: Open-vocabulary 3d scene graphs for perception and planning,'' \emph{arXiv preprint arXiv:2309.16650}, 2023.

\bibitem{vlmaps}
C.~Huang, O.~Mees, A.~Zeng, and W.~Burgard, ``Visual language maps for robot navigation,'' in \emph{2023 IEEE International Conference on Robotics and Automation (ICRA)}.\hskip 1em plus 0.5em minus 0.4em\relax IEEE, 2023, pp. 10\,608--10\,615.

\bibitem{mgpt}
O.~Shliazhko, A.~Fenogenova, M.~Tikhonova, V.~Mikhailov, A.~Kozlova, and T.~Shavrina, ``mgpt: Few-shot learners go multilingual,'' \emph{arXiv preprint arXiv:2204.07580}, 2022.

\bibitem{gato}
S.~Reed, K.~Zolna, E.~Parisotto, S.~G. Colmenarejo, A.~Novikov, G.~Barth-Maron, M.~Gimenez, Y.~Sulsky, J.~Kay, J.~T. Springenberg \emph{et~al.}, ``A generalist agent,'' \emph{arXiv preprint arXiv:2205.06175}, 2022.

\bibitem{crossformer}
R.~Doshi, H.~Walke, O.~Mees, S.~Dasari, and S.~Levine, ``Scaling cross-embodied learning: One policy for manipulation, navigation, locomotion and aviation,'' \emph{arXiv preprint arXiv:2408.11812}, 2024.

\bibitem{robotcat}
K.~Bousmalis, G.~Vezzani, D.~Rao, C.~M. Devin, A.~X. Lee, M.~B. Villalonga, T.~Davchev, Y.~Zhou, A.~Gupta, A.~Raju \emph{et~al.}, ``Robocat: A self-improving generalist agent for robotic manipulation,'' \emph{Transactions on Machine Learning Research}, 2023.

\bibitem{survey}
W.~Zhu, X.~Ma, D.~Ro, H.~Ci, J.~Zhang, J.~Shi, F.~Gao, Q.~Tian, and Y.~Wang, ``Human motion generation: A survey,'' \emph{IEEE Transactions on Pattern Analysis and Machine Intelligence}, 2023.

\bibitem{clip}
A.~Radford, J.~W. Kim, C.~Hallacy, A.~Ramesh, G.~Goh, S.~Agarwal, G.~Sastry, A.~Askell, P.~Mishkin, J.~Clark \emph{et~al.}, ``Learning transferable visual models from natural language supervision,'' in \emph{International conference on machine learning}.\hskip 1em plus 0.5em minus 0.4em\relax PMLR, 2021, pp. 8748--8763.

\bibitem{momask}
C.~Guo, Y.~Mu, M.~G. Javed, S.~Wang, and L.~Cheng, ``Momask: Generative masked modeling of 3d human motions,'' \emph{arXiv preprint arXiv:2312.00063}, 2023.

\bibitem{mdm}
G.~Tevet, S.~Raab, B.~Gordon, Y.~Shafir, D.~Cohen-Or, and A.~H. Bermano, ``Human motion diffusion model,'' \emph{arXiv preprint arXiv:2209.14916}, 2022.

\bibitem{motiondiffuse}
M.~Zhang, Z.~Cai, L.~Pan, F.~Hong, X.~Guo, L.~Yang, and Z.~Liu, ``Motiondiffuse: Text-driven human motion generation with diffusion model,'' \emph{IEEE Transactions on Pattern Analysis and Machine Intelligence}, 2024.

\bibitem{vint}
H.-T.~L. Chiang, Z.~Xu, Z.~Fu, M.~G. Jacob, T.~Zhang, T.-W.~E. Lee, W.~Yu, C.~Schenck, D.~Rendleman, D.~Shah \emph{et~al.}, ``Mobility vla: Multimodal instruction navigation with long-context vlms and topological graphs,'' \emph{arXiv preprint arXiv:2407.07775}, 2024.

\bibitem{gnm}
D.~Shah, A.~Sridhar, A.~Bhorkar, N.~Hirose, and S.~Levine, ``Gnm: A general navigation model to drive any robot,'' in \emph{2023 IEEE International Conference on Robotics and Automation (ICRA)}.\hskip 1em plus 0.5em minus 0.4em\relax IEEE, 2023, pp. 7226--7233.

\bibitem{openx}
O.~X.-E. Collaboration, A.~O'Neill, A.~Rehman, A.~Maddukuri, A.~Gupta, A.~Padalkar, A.~Lee, A.~Pooley, A.~Gupta, A.~Mandlekar, A.~Jain, A.~Tung, A.~Bewley, A.~Herzog, A.~Irpan, A.~Khazatsky, A.~Rai, A.~Gupta, A.~Wang, A.~Kolobov, A.~Singh, A.~Garg, A.~Kembhavi, A.~Xie, A.~Brohan, A.~Raffin, A.~Sharma, A.~Yavary, A.~Jain, A.~Balakrishna, A.~Wahid, B.~Burgess-Limerick, B.~Kim, B.~Schölkopf, B.~Wulfe, B.~Ichter, C.~Lu, C.~Xu, C.~Le, C.~Finn, C.~Wang, C.~Xu, C.~Chi, C.~Huang, C.~Chan, C.~Agia, C.~Pan, C.~Fu, C.~Devin, D.~Xu, D.~Morton, D.~Driess, D.~Chen, D.~Pathak, D.~Shah, D.~Büchler, D.~Jayaraman, D.~Kalashnikov, D.~Sadigh, E.~Johns, E.~Foster, F.~Liu, F.~Ceola, F.~Xia, F.~Zhao, F.~V. Frujeri, F.~Stulp, G.~Zhou, G.~S. Sukhatme, G.~Salhotra, G.~Yan, G.~Feng, G.~Schiavi, G.~Berseth, G.~Kahn, G.~Yang, G.~Wang, H.~Su, H.-S. Fang, H.~Shi, H.~Bao, H.~B. Amor, H.~I. Christensen, H.~Furuta, H.~Walke, H.~Fang, H.~Ha, I.~Mordatch, I.~Radosavovic, I.~Leal, J.~Liang, J.~Abou-Chakra, J.~Kim, J.~Drake, J.~Peters,
  J.~Schneider, J.~Hsu, J.~Bohg, J.~Bingham, J.~Wu, J.~Gao, J.~Hu, J.~Wu, J.~Wu, J.~Sun, J.~Luo, J.~Gu, J.~Tan, J.~Oh, J.~Wu, J.~Lu, J.~Yang, J.~Malik, J.~Silvério, J.~Hejna, J.~Booher, J.~Tompson, J.~Yang, J.~Salvador, J.~J. Lim, J.~Han, K.~Wang, K.~Rao, K.~Pertsch, K.~Hausman, K.~Go, K.~Gopalakrishnan, K.~Goldberg, K.~Byrne, K.~Oslund, K.~Kawaharazuka, K.~Black, K.~Lin, K.~Zhang, K.~Ehsani, K.~Lekkala, K.~Ellis, K.~Rana, K.~Srinivasan, K.~Fang, K.~P. Singh, K.-H. Zeng, K.~Hatch, K.~Hsu, L.~Itti, L.~Y. Chen, L.~Pinto, L.~Fei-Fei, L.~Tan, L.~J. Fan, L.~Ott, L.~Lee, L.~Weihs, M.~Chen, M.~Lepert, M.~Memmel, M.~Tomizuka, M.~Itkina, M.~G. Castro, M.~Spero, M.~Du, M.~Ahn, M.~C. Yip, M.~Zhang, M.~Ding, M.~Heo, M.~K. Srirama, M.~Sharma, M.~J. Kim, N.~Kanazawa, N.~Hansen, N.~Heess, N.~J. Joshi, N.~Suenderhauf, N.~Liu, N.~D. Palo, N.~M.~M. Shafiullah, O.~Mees, O.~Kroemer, O.~Bastani, P.~R. Sanketi, P.~T. Miller, P.~Yin, P.~Wohlhart, P.~Xu, P.~D. Fagan, P.~Mitrano, P.~Sermanet, P.~Abbeel, P.~Sundaresan, Q.~Chen,
  Q.~Vuong, R.~Rafailov, R.~Tian, R.~Doshi, R.~Mart{'i}n-Mart{'i}n, R.~Baijal, R.~Scalise, R.~Hendrix, R.~Lin, R.~Qian, R.~Zhang, R.~Mendonca, R.~Shah, R.~Hoque, R.~Julian, S.~Bustamante, S.~Kirmani, S.~Levine, S.~Lin, S.~Moore, S.~Bahl, S.~Dass, S.~Sonawani, S.~Song, S.~Xu, S.~Haldar, S.~Karamcheti, S.~Adebola, S.~Guist, S.~Nasiriany, S.~Schaal, S.~Welker, S.~Tian, S.~Ramamoorthy, S.~Dasari, S.~Belkhale, S.~Park, S.~Nair, S.~Mirchandani, T.~Osa, T.~Gupta, T.~Harada, T.~Matsushima, T.~Xiao, T.~Kollar, T.~Yu, T.~Ding, T.~Davchev, T.~Z. Zhao, T.~Armstrong, T.~Darrell, T.~Chung, V.~Jain, V.~Vanhoucke, W.~Zhan, W.~Zhou, W.~Burgard, X.~Chen, X.~Chen, X.~Wang, X.~Zhu, X.~Geng, X.~Liu, X.~Liangwei, X.~Li, Y.~Pang, Y.~Lu, Y.~J. Ma, Y.~Kim, Y.~Chebotar, Y.~Zhou, Y.~Zhu, Y.~Wu, Y.~Xu, Y.~Wang, Y.~Bisk, Y.~Dou, Y.~Cho, Y.~Lee, Y.~Cui, Y.~Cao, Y.-H. Wu, Y.~Tang, Y.~Zhu, Y.~Zhang, Y.~Jiang, Y.~Li, Y.~Li, Y.~Iwasawa, Y.~Matsuo, Z.~Ma, Z.~Xu, Z.~J. Cui, Z.~Zhang, Z.~Fu, and Z.~Lin, ``Open {X-E}mbodiment: Robotic learning
  datasets and {RT-X} models,'' \url{https://arxiv.org/abs/2310.08864}, 2023.

\bibitem{diff_nav1}
\BIBentryALTinterwordspacing
D.~Shah, A.~Sridhar, A.~Bhorkar, N.~Hirose, and S.~Levine, ``{GNM: A General Navigation Model to Drive Any Robot},'' in \emph{International Conference on Robotics and Automation (ICRA)}, 2023. [Online]. Available: \url{https://arxiv.org/abs/2210.03370}
\BIBentrySTDinterwordspacing

\bibitem{diff_nav2}
\BIBentryALTinterwordspacing
D.~Shah, A.~Sridhar, N.~Dashora, K.~Stachowicz, K.~Black, N.~Hirose, and S.~Levine, ``Vi{NT}: A foundation model for visual navigation,'' in \emph{7th Annual Conference on Robot Learning}, 2023. [Online]. Available: \url{https://arxiv.org/abs/2306.14846}
\BIBentrySTDinterwordspacing

\bibitem{diff_nav3}
\BIBentryALTinterwordspacing
A.~Sridhar, D.~Shah, C.~Glossop, and S.~Levine, ``{NoMaD: Goal Masked Diffusion Policies for Navigation and Exploration},'' \emph{arXiv pre-print}, 2023. [Online]. Available: \url{https://arxiv.org/abs/2310.xxxx}
\BIBentrySTDinterwordspacing

\bibitem{saytap}
Y.~Tang, W.~Yu, J.~Tan, H.~Zen, A.~Faust, and T.~Harada, ``Saytap: Language to quadrupedal locomotion,'' \emph{arXiv preprint arXiv:2306.07580}, 2023.

\bibitem{AMASS}
N.~Mahmood, N.~Ghorbani, N.~F. Troje, G.~Pons-Moll, and M.~J. Black, ``{AMASS}: Archive of motion capture as surface shapes,'' in \emph{International Conference on Computer Vision}, Oct. 2019, pp. 5442--5451.

\bibitem{humanml}
C.~Guo, S.~Zou, X.~Zuo, S.~Wang, W.~Ji, X.~Li, and L.~Cheng, ``Generating diverse and natural 3d human motions from text,'' in \emph{Proceedings of the IEEE/CVF Conference on Computer Vision and Pattern Recognition (CVPR)}, June 2022, pp. 5152--5161.

\bibitem{t5}
\BIBentryALTinterwordspacing
C.~Raffel, N.~Shazeer, A.~Roberts, K.~Lee, S.~Narang, M.~Matena, Y.~Zhou, W.~Li, and P.~J. Liu, ``Exploring the limits of transfer learning with a unified text-to-text transformer,'' \emph{Journal of Machine Learning Research}, vol.~21, no. 140, pp. 1--67, 2020. [Online]. Available: \url{http://jmlr.org/papers/v21/20-074.html}
\BIBentrySTDinterwordspacing

\bibitem{robot_motion_caption}
T.~Yamada, H.~Matsunaga, and T.~Ogata, ``Paired recurrent autoencoders for bidirectional translation between robot actions and linguistic descriptions,'' \emph{IEEE Robotics and Automation Letters}, vol.~3, no.~4, pp. 3441--3448, 2018.

\bibitem{vqvae}
A.~Razavi, A.~Van~den Oord, and O.~Vinyals, ``Generating diverse high-fidelity images with vq-vae-2,'' \emph{Advances in neural information processing systems}, vol.~32, 2019.

\bibitem{wordpiece}
X.~Song, A.~Salcianu, Y.~Song, D.~Dopson, and D.~Zhou, ``Fast wordpiece tokenization,'' \emph{arXiv preprint arXiv:2012.15524}, 2020.

\bibitem{SMPL}
M.~Loper, N.~Mahmood, J.~Romero, G.~Pons-Moll, and M.~J. Black, ``{SMPL}: A skinned multi-person linear model,'' \emph{ACM Trans. Graphics (Proc. SIGGRAPH Asia)}, vol.~34, no.~6, pp. 248:1--248:16, Oct. 2015.

\bibitem{ppo}
J.~Schulman, F.~Wolski, P.~Dhariwal, A.~Radford, and O.~Klimov, ``Proximal policy optimization algorithms,'' \emph{arXiv preprint arXiv:1707.06347}, 2017.

\bibitem{instruction}
S.~Zhang, L.~Dong, X.~Li, S.~Zhang, X.~Sun, S.~Wang, J.~Li, R.~Hu, T.~Zhang, F.~Wu \emph{et~al.}, ``Instruction tuning for large language models: A survey,'' \emph{arXiv preprint arXiv:2308.10792}, 2023.

\bibitem{BABEL}
A.~R. Punnakkal, A.~Chandrasekaran, N.~Athanasiou, A.~Quiros-Ramirez, and M.~J. Black, ``{BABEL}: Bodies, action and behavior with english labels,'' in \emph{Proceedings IEEE/CVF Conf.~on Computer Vision and Pattern Recognition (CVPR)}, Jun. 2021, pp. 722--731.

\bibitem{gpt3}
T.~B. Brown, ``Language models are few-shot learners,'' \emph{arXiv preprint arXiv:2005.14165}, 2020.

\bibitem{diederik2014adam}
P.~K. Diederik, ``Adam: A method for stochastic optimization,'' \emph{(No Title)}, 2014.

\bibitem{FID}
H.~Martin, R.~Hubert, U.~Thomas, N.~Bernhard, and H.~Sepp, ``Gans trained by a two time-scale update rule converge to a local nash equilibrium,'' \emph{Advances in neural information processing systems}, vol.~30, pp. 6626--6637, 2017.

\bibitem{bertscore}
T.~Zhang, V.~Kishore, F.~Wu, K.~Q. Weinberger, and Y.~Artzi, ``Bertscore: Evaluating text generation with bert,'' \emph{arXiv preprint arXiv:1904.09675}, 2019.

\bibitem{rouge}
L.-Y. ROUGE, ``A packageforautomaticevaluationof summaries,'' \emph{ProcofPost-ConferenceWorkshoponText SummarizationBranchesOutofthe42nd AnnualMeeting onAssociationforComputationalLinguistics}, pp. 74--81, 2004.

\bibitem{cider}
R.~Vedantam, C.~Lawrence~Zitnick, and D.~Parikh, ``Cider: Consensus-based image description evaluation,'' in \emph{Proceedings of the IEEE conference on computer vision and pattern recognition}, 2015, pp. 4566--4575.

\bibitem{bleu}
K.~Papineni, S.~Roukos, T.~Ward, and W.~Zhu, ``A method for automatic evaluation of machine translation'','' \emph{the Proceedings of ACL-2002, ACL, Philadelphia, PA, July 2002}, 2001.

\bibitem{diffuser}
M.~Janner, Y.~Du, J.~Tenenbaum, and S.~Levine, ``Planning with diffusion for flexible behavior synthesis,'' in \emph{International Conference on Machine Learning}, 2022.

\end{thebibliography}
}

\end{document}